\documentclass[letterpaper, 10 pt, conference]{ieeeconf}  

\IEEEoverridecommandlockouts                              
\usepackage{graphics} 
\usepackage{caption}
\usepackage[ruled,vlined]{algorithm2e}
\usepackage{epsfig} 
\usepackage{mathptmx} 
\usepackage{times} 
\usepackage{amsmath} 
\usepackage{subfig}
\usepackage{float}
\usepackage{amssymb}  
\usepackage{bbm}
\usepackage{url}
\usepackage[utf8]{inputenc}
\usepackage[english]{babel}
\usepackage{color}

\renewcommand{\baselinestretch}{0.91}
\allowdisplaybreaks

\DeclareMathOperator*{\argmin}{arg\,min}
\DeclareMathOperator*{\argmax}{arg\,max}

\newcommand{\xr}{x_\mathrm{r}}
\newcommand{\ur}{u_\mathrm{r}}
\newcommand{\xo}{x_\mathrm{o}}
\newcommand{\uo}{u_\mathrm{o}}
\newcommand{\zr}{z_\mathrm{r}}
\newcommand{\wwr}{w_\mathrm{r}}
\newcommand{\zo}{z_\mathrm{o}}
\newcommand{\wo}{w_\mathrm{o}}
\newcommand{\xrt}[1]{x_\mathrm{r}^{(#1)}}
\newcommand{\urt}[1]{u_\mathrm{r}^{(#1)}}
\newcommand{\zrt}[1]{z_\mathrm{r}^{(#1)}}
\newcommand{\wrt}[1]{w_\mathrm{r}^{(#1)}}
\newcommand{\zot}[1]{z_\mathrm{o}^{(#1)}}
\newcommand{\wot}[1]{w_\mathrm{o}^{(#1)}}
\newcommand{\xot}[1]{x_\mathrm{o}^{(#1)}}
\newcommand{\uot}[1]{u_\mathrm{o}^{(#1)}}

\newcommand{\xrel}{x_\mathrm{rel}}
\newcommand{\xrelt}[1]{x_\mathrm{rel}^{(#1)}}

\newcommand{\HJI}{\mathrm{HJI}}

\newcommand{\pxr}{p_{x,\mathrm{r}}}
\newcommand{\pyr}{p_{y,\mathrm{r}}}
\newcommand{\vr}{v_\mathrm{r}}
\newcommand{\thetar}{\theta_\mathrm{r}}
\newcommand{\omr}{\omega_\mathrm{r}}
\newcommand{\ar}{a_\mathrm{r}}

\newcommand{\pxo}{p_{x,\mathrm{o}}}
\newcommand{\pyo}{p_{y,\mathrm{o}}}
\newcommand{\vo}{v_\mathrm{o}}

\newcommand{\thetao}{\theta_\mathrm{o}}

\newcommand{\ao}{a_\mathrm{o}}

\newcommand{\pxrel}{p_{x,\mathrm{rel}}}
\newcommand{\pyrel}{p_{y,\mathrm{rel}}}

\newcommand{\px}{p_{x}}
\newcommand{\py}{p_{y}}

\newcommand{\musafe}{\mu_{\mathrm{safe}}}

\title{\LARGE \bf
Infusing Reachability-Based Safety into\\Planning and Control for Multi-agent Interactions
}

\author{Xinrui Wang$^{1\star}$, Karen Leung$^{2\star}$, Marco Pavone$^{2}$
\thanks{$^1$Department of Mechanical Engineering, Stanford University, Stanford, CA 94305. {\tt\small xinrui@stanford.edu}}%
\thanks{$^2$Department of Aeronautics and Astronautics, Stanford University, Stanford, CA 94305. {\tt\small \{karenl7, pavone\}@stanford.edu}}%
\thanks{$^\star$ Indicates equal contribution}%
\thanks{This work was supported by the Office of Naval Research (Grant N00014- 17-1-2433) and by the Toyota Research Institute (“TRI”). This article solely reflects the opinions and conclusions of its authors and not ONR, TRI or any other Toyota entity.}
}

\begin{document}

\maketitle
\thispagestyle{empty}
\pagestyle{empty}

\begin{abstract}
Within a robot autonomy stack, the planner and controller are typically designed separately, and serve different purposes. As such, there is often a diffusion of responsibilities when it comes to ensuring safety for the robot. 
We propose that a planner and controller should share the same interpretation of safety but apply this knowledge in a different yet complementary way.
To achieve this, we use Hamilton-Jacobi (HJ) reachability theory at the planning level to provide the robot planner with the foresight to avoid entering regions with possible inevitable collision. However, this alone does not guarantee safety. In conjunction with this HJ reachability-infused planner, we propose a minimally-interventional multi-agent safety-preserving controller also derived via HJ-reachability theory. The safety controller maintains safety for the robot without unduly impacting planner performance.
We demonstrate the benefits of our proposed approach in a multi-agent highway scenario where a robot car is rewarded to navigate through traffic as fast as possible, and we show that our approach provides strong safety assurances yet achieves the highest performance compared to other safety controllers.
\end{abstract}



\section{Introduction}
Decision-making and control for robots is typically stratified into levels, with each having different purposes. 
The high-level planner, informed by representative yet simplified dynamics of a robot and its environment, is designed to be far-sighted and selects plans that optimize performance metrics (e.g., minimize time to goal, control effort, and perception uncertainty). While the low-level controller, running at a much higher frequency than the planner, tends to be more short-sighted and respects more accurate models of the robot's dynamics and control constraints in order to implement the controls necessary to follow the desired plan.
However, when it comes to ensuring safety for the robot, the divide between these components can lead to a diffusion of responsibilities---the planner and controller may each devise their own safety protocols, or even assume the other bears the full responsibility, but when combined together, they may not necessarily complement each other in achieving the shared goal of ensuring safety for the system.

\begin{figure}[t]
    \centering
    \includegraphics[width=0.48\textwidth]{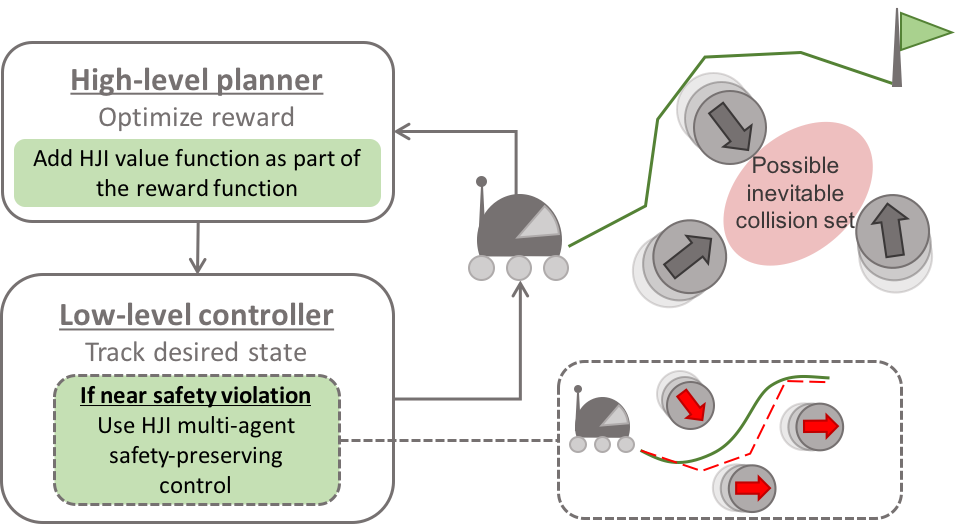}
    \caption{Both the planner and controller use HJ reachability theory to measure safety of the system, but apply this knowledge in complementary ways.
    The high-level planner has the foresight to avoid regions of possible inevitable collision.
    When near safety violation, the safety controller evades multiple agents with minimal intervention (green trajectory), as opposed to a reactive controller (red trajectory).}
    \label{fig:schematic}
\end{figure}

Safety considerations at the planning level can help discourage a robot from entering potentially dangerous situations.
However, ensuring safety typically competes with other planning objectives (e.g., minimizing time and control effort). Additionally, simplifying assumptions that underpin a planner's model of the environment may cause any strong safety assurances claimed by the planner to no longer carry much weight in practice due to model mismatch with the real system.
A low-level controller, whose primary purpose is to track the desired plan, is typically too shortsighted to include safety considerations especially with respect to dynamic obstacles. Instead, the controller may switch to a safety controller when near safety violation. The safety controller may be an augmentation of the existing low-level controller, or a different, possibly lower-level, reactive control scheme.
However, a switching safety controller that completely overrides the system may severely compromise planner performance and potentially create new unsafe situations (e.g., an autonomous vehicle stopping abruptly on the highway can cause rear-end collisions).

In this work, we design a safe complementary planning and control stack that provides safety assurance for a robot without unduly impacting planner performance. We leverage Hamilton-Jacobi (HJ) reachability theory to provide the same representation of safety for both the planner and safety controller. In particular, the planner and safety controller maintain their unique purposes, but share the same sentiment when it comes to understanding safety.
Further, in a multi-agent setting, a robot may be in conflict as to which agent to avoid if there is a danger of colliding with multiple agents. We use HJ reachability theory to optimally select safe controls that reason about collision avoidance with all agents threatening the robot's safety in a minimally interventional way.
We demonstrate the efficacy of the proposed framework in a multi-agent highway setting where an autonomous car must move quickly and safely through a densely populated highway. 
The case study shows that by the planner and controller
sharing their interpretation of safety, the overall interaction yields safe interactions that are more efficient and higher performing compared to reactive safety controllers.

\emph{Related work:} 
Safety at the planning level can be enforced via hard constraints or incentivized through a planner's objective function. 
A common approach is to select plans that avoid the inevitable collision set (ICS), though in practice due to computational tractability, the forward reachable set is used instead which is an over-approximation.
For example, \cite{AlthoffDolan2014,LiuRoehmEtAl2017} prevent the robot's planned trajectory from entering the forward reachable set of the other agents in the environment and ensure the existence of a safe stopping maneuver at all times. 
When no feasible trajectories can be found, the robot switches to the emergency maneuver computed from the last feasible plan.
While these approaches guarantee safety with respect to their modeling assumptions, they tend to be overly-conservative for interactive scenarios, and rely on strong modeling assumptions of other agents.

Alternatively, consideration of interaction dynamics between agents can reduce conservatism and enable proactive behaviors. In the context of autonomous driving, a probabilistic prediction model of the environment can be learned from data and then used to inform a robot's decision making algorithm which strives to optimize a multi-objective function with safety being one of the objectives \cite{SchmerlingLeungEtAl2018,SadighSastryEtAl2016c}. 
This can lead to more efficient but potentially unsafe interactions because safety competes with other objectives.
\cite{MaloneChiangEtAl2017} designs artificial potential fields (APF) that reflect the reachable sets of the other dynamic agents and use it for path planning. They show that this improves the safety of a robot compared to traditional gaussian APF methods.
Regardless of how safety is enforced at the planning level, model mismatch and stochasticity in the environment make it very challenging to provide strict safety assurance.
\begin{figure}[t]
    \centering
    \includegraphics[width=0.37\textwidth]{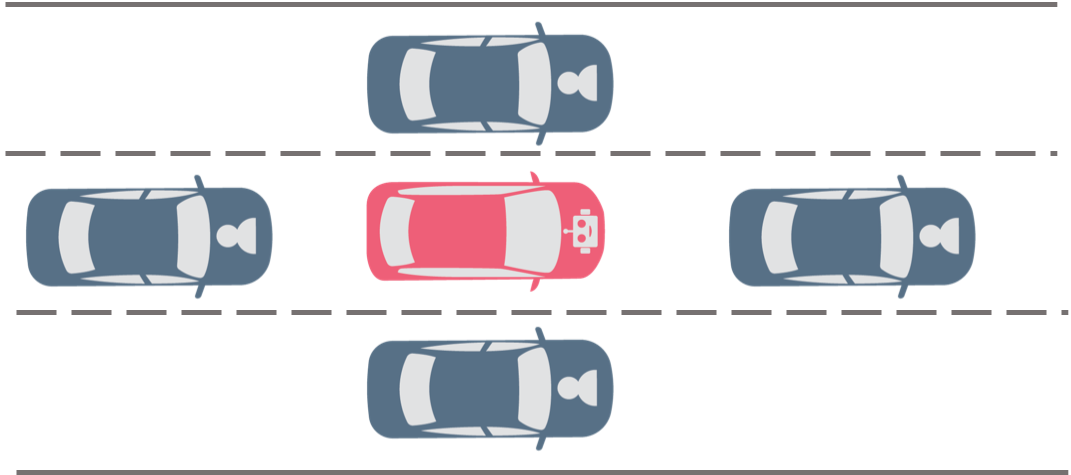}
    \caption{A robot car (red) boxed-in by other externally controlled (e.g., human-driven) cars (blue). If all cars behave adversarially, collision is inevitable.}
    \label{fig:boxed in}
\end{figure}

In turn, low-level controllers typically provide safety reactively; they switch to a safety controller when the system is near safety violation. For instance, \cite{AroraChoudhuryEtAl2015} selects a maneuver from a library of pre-computed emergency maneuvers when the system is in an unsafe state. A paradigm that can account for general interactive scenarios is HJ reachability analysis; by formulating the interactions between a robot and its environment (e.g., other agents) as an optimal control problem, a collision avoidance control can be computed and applied when near safety violation \cite{BajcsyBansalEtAl2019,ChenTomlin2018,FisacAkametaluEtAl2018}.
However, a switching-based safety controller that completely overrides the system may not be ideal in some contexts, such as in autonomous driving where sudden reactions could disrupt traffic flow.
Accordingly, \cite{LeungSchmerlingEtAl2019} proposes a less invasive approach in passing a HJ reachability-based safety-preserving control constraint into a robot's low-level tracking controller. This results in a minimally interventional safety controller that enables a system to deviate from the desired trajectory to the extent necessary to maintain safety. Through a traffic-weaving case study with a single robot car interacting with a single human-controlled car, they show that their approach provides a good balance between safety and efficiency---the robot stays safe without unduly impacting planner performance.
In the same vein, approaches that use control velocity obstacles (CVO) (e.g., \cite{BestNarangEtAl2017}) consider the set of robot controls that prevent collisions with other agents in the future. However, CVO requires an assumption on the other agent's policy which is difficult to obtain for interactive and stochastic settings, and reasons about the future in an open-loop fashion. This is in contrast to HJ reachability, which reasons about collision-avoidance controls under closed-loop dynamics.

Rather than treating the planning and control problem separately, \cite{WurtsSteinEtAl2018,Laurense2019} combine the two into a single joint optimization problem. They show through an aggressive lane change maneuver and an autonomous car racing example that combining planning and control can be more desirable because the resulting trajectory respects more accurate dynamic capabilities of the car while also cognizant of safety constraints such as avoiding static obstacles and staying within road boundaries. 

The key research goal of this paper is to bring the planning and control modules closer together in order to provide stronger safety assurances in the context of multi-agent interactions as illustrated in Figure~\ref{fig:schematic}.
Solely relying on a safety controller may be insufficient because an optimal collision avoidance control may not always exist when near safety violation, especially in the case of multi-agent scenarios. 
For instance, consider an autonomous car that is boxed in, as shown in Figure~\ref{fig:boxed in}; in this situation, a collision is imminent if any of the adjacent cars swerve towards it.
This necessitates the need to jointly design a planner and safety controller---the planner should be cognizant of what type of situations the safety controller is able to succeed in, while the safety controller should be aware of the planner's objectives such that it does not unduly impact overall performance.
 
                                                               
\emph{Statement of Contributions:}
The key contribution of this paper is in using HJ reachability theory to provide a shared notion of safety between the planner and controller of an autonomy stack. Specifically, the contributions are threefold.
First, we provide high-level planners that optimize an objective function with the foresight to avoid regions of possible inevitable collision by incorporating a term into the objective function derived from reachability theory. 
Second, when the system is near safety violation, we propose a multi-agent safety-preserving controller that is minimally interventional---the controller optimizes for performance while treating safety as a constraint. This multi-agent safety controller builds upon the work in \cite{LeungSchmerlingEtAl2019} by extending the controller to multi-agent settings.
Third, we demonstrate our approach in a highway scenario where an autonomous car must safely traverse the highway as fast as possible. 
We perform an ablation study and compare against a reactive control strategy, and a baseline safety controller. We show that our proposed approach offers a good balance between safety and performance---we score relatively high in both performance and safety metrics. The insights gained from the experimental results are general since our approach applies to any planner-control framework as long as HJ reachability analysis is applicable to the system. 

\emph{Organization:}
The remainder of the paper is organized as follows. 
Section \ref{sec:problem formulation} introduces the problem we are approaching and Section \ref{sec: HJ Reachability Analysis} provides a concise overview of HJ reachability analysis. 
Section \ref{sec: align planner and controller} describes our proposed methodology of aligning the planner and safety controller by providing a shared notion of safety via HJ reachability analysis. 
In Section~\ref{sec: case study}, we illustrate the benefits of our approach with a case study in a multi-agent highway driving scenario. 
We conclude in Section \ref{sec: conclusion} and suggest several directions for future work.


\section{Problem Formulation}\label{sec:problem formulation}
We follow a typical decision making and control paradigm of utilizing a high level planner to compute a coarse trajectory and a low-level controller to track it.

\subsection{Robot Planner for Interactive Multi-agent Scenarios}\label{subsec:problem formulation - planner}
We consider a multi-agent setting where a robot is operating in an environment with multiple agents not controlled by the robot (e.g., humans).
Let $\xrt{t}\in\mathcal{X}\subset\mathbb{R}^{n_\mathrm{r}}$ and $\urt{t}\in\mathcal{U}\subset\mathbb{R}^{m_\mathrm{r}}$ be the robot planner state and action at time step $t$ respectively. 
Let $\xot{t}\in\mathcal{X}_\mathrm{o}\subset\mathbb{R}^{Jn_\mathrm{o}}$ and $\uot{t}\in\mathcal{U}_\mathrm{o}\subset\mathbb{R}^{Jm_\mathrm{o}}$ be the state and control of $J$ other agents in the environment at time step $t$. 
Further, let the time-invariant and discrete-time state space dynamics for a robot and other agents in the environment be given respectively by, $\xrt{t+1} = f_\mathrm{r}(\xrt{t},\, \urt{t}), \: \xot{t+1}= f_\mathrm{o}(\xot{t},\, \uot{t}).
$
The goal of a robot planner is to find a sequence of robot actions $\urt{t:t+N} = \pi(\xrt{t}, \,\xot{t})$ that maximizes an expected reward $R(\xrt{t},\, \urt{t},\, \xot{t})$ over a fixed horizon of length $N$. That is, we want to find a solution to the following maximization problem
\begin{align}
    \pi^*(\xrt{t},\, \xot{t}) = \argmax_{\urt{t:t+N}} \: \mathbb{E}\left[ \sum_{i=0}^N \gamma^i R(\xrt{t+i}, \,\urt{t+i},\, \xot{t+i}) \right]
    \label{eqn:planner policy}
\end{align}
where $\gamma \in [0, 1]$ is a discount factor. 
Note that for interactive scenarios, \eqref{eqn:planner policy} is generally difficult to solve due to stochasticity and coupling in the dynamics between the robot and other agents, and that system may not be Markovian (i.e., depend on interaction history) \cite{SadighSastryEtAl2016c,LeungSchmerlingEtAl2019}. 
We assume that such planners typically operate at around $<$10 Hz. As such, they are not always able to account for split-second threats, necessitating the need for a safety controller to intervene.

\subsection{Low-level Safety Controller}\label{subsec:problem formulation - controller}
A low-level controller, operating at a higher frequency than the planner, takes the desired trajectory produced by the planner (computed by simulating the system with the desired action sequence) and computes relevant low-level control commands for the system to execute. More concretely, let $\zrt{t}\in\mathcal{Z}\subset \mathbb{R}^{p_\mathrm{r}}$ and $\wrt{t}\in\mathcal{W}\subset\mathbb{R}^{q_\mathrm{r}}$ represent the low-level state and control at time step $t$ respectively. The low-level controller may use a higher fidelity model than the planner, $\zrt{t+1} = f_\mathrm{c}(\zrt{t},\, \wrt{t}),$
in order to compute appropriate actuation commands for the system. 
Then $\mu:\mathcal{Z} \times \mathcal{X} \rightarrow \mathcal{W}: \mu(\zrt{t},\xr) \mapsto \wrt{t}$ is a policy that maps current controller state and desired planner state to actuation commands. 
When near safety violation, the system may switch to a reactive policy $\musafe$ that selects control actions that keep the system safe with respect to some metric $C_\mathrm{safe}(\zr,\, \wwr,\, \zo)$ where $\zo$ is the low-level state of the other agents in the environment. That is, in the most general sense, the optimal safe policy $\musafe^*$ is a solution to the following optimization problem,
\begin{equation*}
\begin{aligned}
    \argmin_{\wwr\in\mathcal{W}}\;C_\mathrm{safe}(\zr,\, \wwr,\, \zo), \quad
    \mathrm{s.t.} \;\;\  \text{System is safe} .
    \label{eqn:safety control policy}
\end{aligned}
\end{equation*}
which selects the safest action. Slack variables can be used to ensure feasibility of the problem.


\section{Hamilton-Jacobi Reachability Analysis}
\label{sec: HJ Reachability Analysis}
In this section, we highlight key HJ backward reachability concepts relevant to our proposed planning and control strategy; see \cite{ChenTomlin2018} for a more in-depth overview.

\subsection{Overview}
Given a dynamics model governing a robotic system incorporating control and disturbance inputs, reachability analysis is the study of the set of states that the system can reach from its initial conditions, i.e., the reachable set.
It is often used for formal verification as it can give guarantees on whether or not the evolution of the system will be unsafe, i.e., whether the reachable set includes undesirable outcomes. 
In this work, we use \emph{backward} reachability analysis because (i) of its non-overly conservative nature stemming from closed-loop computations, and (ii) the set is computed offline and provide near-instant access online via table look-up.
See \cite{LeungSchmerlingEtAl2019} for a more in-depth discussion on the suitability of backward reachability for interactive scenarios.

There are many existing approaches to compute the reachable set of a system \cite{GreenstreetMitchell1998,KurzhanskiVaraiya2000,FrehseLeGuernicEtAl2011,AlthoffKrogh2014,MajumdarVasudevanEtAl2014}, but there is a trade-off among modeling assumptions, scalability, and representation fidelity. Compared to alternative approaches, HJ reachability uses a brute force computation via dynamic programming and thus is the most computationally expensive. However, HJ reachability is able to compute the reachable set exactly\footnote{With precision dependent on parameters of the numerical solver, e.g., discretization choices in mesh size/time step.} for any general nonlinear dynamics with control and disturbance inputs. Further, since the sets are computed offline, we can access the set information online via a near-instant look-up therefore enabling high operating frequencies.
\subsection{Backward Reachable Tube}
In this section, we suppose that a system has
dynamics $\dot{x} = f(x, u, d)$ where $x\in\mathbb{R}^n$ is the state, $u\in\mathcal{U} \subset \mathbb{R}^m$ is the control, and $d\in\mathcal{D} \subset\mathbb{R}^p$ is the disturbance. 
The system dynamics $f:\mathbb{R}^n \times \mathcal{U} \times \mathcal{D} \rightarrow \mathbb{R}^n$ are assumed to be uniformly continuous, bounded, and Lipschitz continuous in $x$ for a fixed $u$ and $d$. 
Let $\mathcal{T} \subseteq \mathbb{R}^n$ be the target set that the system wants to avoid.
For collision avoidance, $\mathcal{T}$ typically represents the set of states that are in collision with an obstacle. 
The \emph{backward reachable tube} (BRT) is the set of states that could result in the system entering the target set under worst-case disturbances within some time horizon $|\tau|$ ($\tau$ is negative since we are propagating backwards in time).
The BRT is denoted by $\mathcal{A}(\tau)$ and is defined as
\begin{equation*}
\begin{split}
    &\mathcal{A}(\tau)  := \lbrace  \bar{x} \in \mathbb{R}^n: \exists d(\cdot), \forall u (\cdot), \exists s\in[\tau, 0], \\
    & \qquad \qquad (x(\tau) = \bar{x})\:\wedge\: (\dot{x} = f(x, u, d)) \:  \wedge\: ( x(s) \in \mathcal{T}) \rbrace.
\end{split}
    \label{eqn: backward reachable tube}
\end{equation*}
$\mathcal{A}(\tau)$ represents the set of states from which there does not exist a policy for the system that can prevent the dynamics from being driven into the target set under worst-case disturbances within a time horizon $|\tau|$.
As such, to rule out such an eventuality, the BRT is treated as the ``avoid set.''

                                                                   
\subsection{HJI Value Function }
Assuming optimal (i.e., adversarial) disturbances, $\mathcal{A}(\tau)$ can be computed by defining a value function $V(\tau, x)$ which obeys the Hamilton-Jacobi-Issacs (HJI) partial differential equation (PDE) \cite{MitchellBayenEtAl2005}; the solution $V(\tau, x)$ gives the BRT as its zero sublevel set, 
\begin{align*}
\mathcal{A}(\tau) = \lbrace x : V(\tau,x) \leq 0\rbrace.
\end{align*}
The HJI PDE is solved starting from the boundary condition $V(0, x)$, the sign of which reflects set membership of $x$ in $\mathcal{T}$. We cache the solution $V(\tau,x)$ to be used online as a look-up table.
For collision avoidance, $V(0,x)$ typically represents the signed distance between a robot and an obstacle ($x\in\mathcal{T} \iff V(0,x)\leq 0$). 
Then the solution $V(\tau,x)$ represents the minimum signed distance between the robot and the obstacle within a time horizon $|\tau|$ if the robot follows an optimal policy under worst-case (i.e., adversarial) disturbances.
As such, the value function can be interpreted as a quantitative measure of robot safety; the larger the value, the safer.


\subsection{HJI Optimal Control}
Given the HJI value function $V(\tau,x)$, the optimal robot policy under worst-case disturbances is,
\begin{align}
    u^* = \mathrm{arg}\max_u \min_d \nabla V(\tau,x)^T f(x,u,d).
    \label{eqn:HJI optimal control}
\end{align}
Many robotic systems employing HJ reachability-based safety switch to using \eqref{eqn:HJI optimal control} when near safety violation (i.e., when $V(t,x) \leq \epsilon,\, \epsilon > 0$) \cite{BajcsyBansalEtAl2019,ChenTomlin2018,FisacAkametaluEtAl2018}.
Recently, \cite{LeungSchmerlingEtAl2019} considers a less invasive control strategy and computes the set of safety-preserving controls,
\begin{align}
    \mathcal{U}_\mathrm{safe}(x) = \{ u\in\mathcal{U} \mid \min_d \nabla V(\tau,x)^T f(x,u,d) \geq 0\}
    \label{eqn:safety preserving}
\end{align}
which describes the set of controls that keep the value function from decreasing under worst-case disturbances. By passing $\mathcal{U}_\mathrm{safe}$ as a control constraint over the next time step when computing appropriate low-level controls, the system is able to minimally deviate from the desired plan to the extent necessary to maintain safety.

\subsection{Collision Avoidance With Multiple Dynamic Agents}\label{subsec:HJI - multiple agents}
We can consider collision avoidance between two dynamic agents by letting $x$ represent the relative state of the system, $u$ correspond to the control inputs of the agent to be controlled (i.e., the robot), and $d$ correspond to the control inputs of the other agent.
The target set $\mathcal{T}$ is the set of relative states corresponding to the system being in an undesirable situation (e.g., in collision).
In theory it is also possible to formulate the relative state for more than two agents, but is impractical because HJ reachability suffers from the curse of dimensionality.
We treat the pairwise interactions independently (i.e., ignore interactions between other agents) to circumvent the scalability issue. However, this treatment can lead to issues regarding which pairwise interaction to prioritize first. 
We address the prioritization issue in this work by proposing an optimization problem that considers all violating pairs equally.


\section{Aligning the Planner and Controller}
\label{sec: align planner and controller}
\subsection{Overview}
Although the purpose of a safety controller is to keep the robot safe when near safety violation, it is possible that the robot may enter a state of possible inevitable collision, due to shortcomings of the planner.
Even when maintaining safety is feasible, the resulting safety control may go against the direction of the planner and lead to chattering.
We propose using HJ reachability theory to create alignment between a planner and safety controller by ensuring that they both share the same representation of safety.
Further, in a multi-agent scenario, a robot safety controller may be in conflict when there are multiple agents to avoid.
We use HJ reachability theory to analyse pairwise safety (see Section~\ref{subsec:HJI - multiple agents}) and propose a computationally efficient solution that prioritizes collision avoidance with all pairs equally. 

\subsection{Assumptions}
We assume control over one autonomous agent (i.e., a robot) in an environment with multiple agents uncontrolled by the robot (e.g., humans). 
    In addition to the problem formulation introduced in Section~\ref{sec:problem formulation}, we make the following assumptions.
Suppose there are $J$ other agents in the environment. Let $\xrelt{j}$ denote the relative state between the robot and the $j$th agent. The relative state is derived from the robot's low-level state $\zr$ and the agent's low-level state $\zot{j}$. The robot control is $\wwr$ and the other agent's control is $\wot{j}$. 
The HJI value function corresponding to the $j$th pairwise system is denoted by $V^{(j)}(\tau, \xrelt{j})$.

\subsection{HJI-aware Interaction Planner}
A high-level planner selects actions that maximize a reward shown in \eqref{eqn:planner policy} which reflects desired goals, such as reaching a goal state, maintaining speed, or reducing time. 
We propose adding a term encompassing the HJI value function between the robot and all other agents into a planner's reward function,
\begin{align}
    R_\mathrm{total}(\xr, \,\ur, \,\xo, \xrel) = \gamma_R R(\xr,\, \ur,\, \xo) + (1 -\gamma_R) R_\HJI(\xrel),
    \label{eqn: HJI planner reward function}
\end{align}
where $\xrel$ is the relative state between the robot and all other agents.
The designer can choose how to define $R_\HJI(\xrel)$ depending on the application. For example, we use $R_\HJI(\xrel) =  \min_{j} \; V^{(j)}(\tau, \xrel^{(j)})$
in the highway driving scenario studied in this paper.
The designer can adjust $\gamma_R$ to tune the robot's preference for reward-seeking or safety-seeking behavior.
When the target set $\mathcal{T}$ is the set of collision states (or a proxy for safety), the value function represents an upper bound on how safe the future will be under worst-case disturbances. 
With all else equal, a planner with an additional HJI term in its reward function will select plans that are safer with respect to backward reachability theory, i.e., the planner will avoid states of possible inevitable collision, such as being boxed-in by other agents (see Figure~\ref{fig:boxed in}).

We make little assumptions on the structure of the planner---the planner could reason about the interaction dynamics jointly unlike our pairwise HJ reachability formulation. The only assumption we make is that the planner strives to maximize a reward function, and can accommodate the addition of the proposed HJI term. For instance, this could be applied to the planner used in \cite{SchmerlingLeungEtAl2018}.

                                       
\subsection{Multi-agent Safety-Preserving Control}
A system may fall back to a safety controller when near safety violation. In HJ reachability applications, this occurs when $V(\tau,x)\leq \epsilon,\: \epsilon > 0$, i.e., when the system is close to the boundary of the BRT. 
Since we consider pairwise interactions between a robot and every other agent in the environment, there is the problem of prioritizing which pairwise interaction should be tackled first if safety is nearly violated for multiple pairs. 
To address this issue, we propose selecting feasible controls from the intersection of the safety-preserving control sets \eqref{eqn:safety preserving} of all pairwise systems where safety is nearly violated. 
Let
\[\mathcal{K} = \{ j \mid V^{(j)}(\tau, \xrelt{j}) \leq \epsilon \quad \text{for} \: j = 1,\,\ldots,\, J\}\]
be the set of indices $k$ where safety is violated for that robot-agent pair.
Further, we strive to select controls that are minimally interventional---controls that maintain safety of the system without large deviations from the nominal controls. As such, we propose the optimal safety controller to be the solution of the optimization problem,
\begin{equation}
\begin{aligned}
    \min_{\wwr}  \;\; g(\wwr),\quad 
    \mathrm{s.t.}  \;\; \wwr \in \bigcap_{k\in \mathcal{K}} \mathcal{U}_\mathrm{safe}^{(k)}(\xrel^{(k)})\: \wedge \: \wwr \in \mathcal{W}
    \label{eqn:HJI multiagent safety controller}
\end{aligned}
\end{equation}
where $g$ is a cost function on $\wwr$ which may strive to achieve high tracking performance or minimize control effort.
In the case where a feasible solution to \eqref{eqn:HJI multiagent safety controller} exists, HJ reachability theory guarantees that the system will not enter a less safe state with respect to the chosen dynamics, i.e., the value function will not decrease.
For control affine systems, $ \mathcal{U}_\mathrm{safe}^{(k)}(\xrel^{(k)})$ is a hyperplane.
In general, \eqref{eqn:HJI multiagent safety controller} could be nonlinear and intractable to solve, especially at a high operating frequency.
Techniques to convexify the problem can be applied to make \eqref{eqn:HJI multiagent safety controller} tractable (e.g., \cite{LeungSchmerlingEtAl2019}). However, the problem will no longer retain strict safety guarantees afforded by HJ reachability theory. Slack variables can be used to ensure the problem remains feasible and that the least-violating control is chosen.


\section{Case Study: Autonomous Highway Driving}
\label{sec: case study}
\begin{figure}[t]
    \centering
    \includegraphics[width=\linewidth]{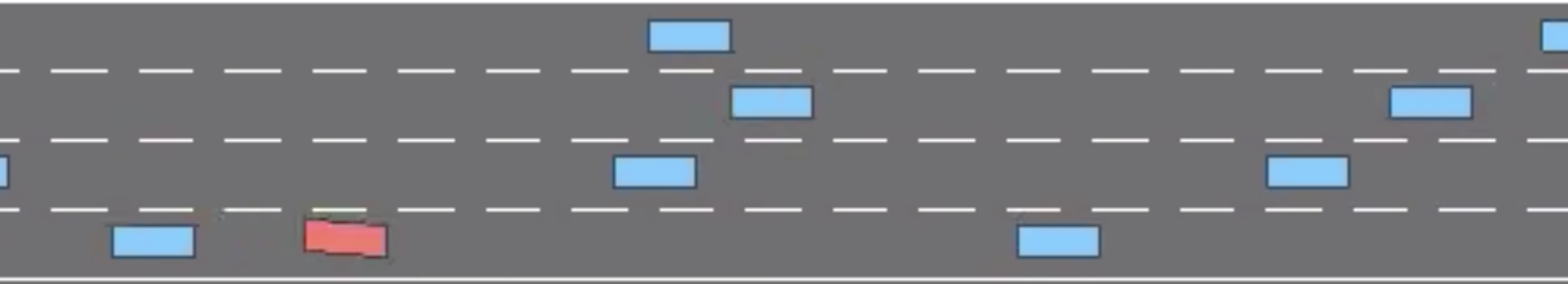}
    \caption{Snapshot from the highway simulation environment. The robot car (red) is tasked to drive as fast as possible through the traffic but avoid collision with other cars (blue).}
    \label{fig:highway env}
\end{figure}

\subsection{Problem Set-up}
Our experiments are conducted in a highway simulation environment (see Figure~\ref{fig:highway env}) developed by \cite{Leurent2018}. The robot car is tasked to drive through traffic as fast as possible while avoiding collision.
The other cars on the road interact with each other using the Intelligent Driver Model (IDM) \cite{TreiberHenneckeEtAl2000,TreiberHenneckeEtAl2000b} and minimizing-overall-braking-induced-by-lane-change (MOBIL) model \cite{KestingTreiberEtAl2007} for longitudinal and lateral control, respectively. These are common, and well-studied traffic flow models.
Our results, however, do not depend critically on this modeling choice.
The high level planner runs at 1 Hz while the low level controller runs at 50 Hz.

\subsubsection{High-level Planner}\label{subsubsec:case study high level planner}
We formulate the highway environment as a Markov decision process (MDP) and apply the optimistic planning (OP) algorithm proposed in \cite{HrenMunos2008}. OP is a tree search algorithm where each branch represents a possible future and the tree is explored optimistically.
We omit implementation details about the planner since our proposed approach is agnostic to the type of planner used. 
Instead we highlight some key features regarding our planner that is representative of the class of planners that our approach is applicable to.
First, our planner relies on a model of the environment, but in general, this model is only an approximation of the true system. 
The existence of model mismatch is prevalent in many model-based control problems, including those for stochastic environments.
In this case study, the robot does not have access to the modeling parameters of the other cars which are drawn from a normal distribution, but instead assumes the mean of the distribution. 
Second, our planner is reward-based and safety is promoted via the objective function. 
As such, there will be times where the robot will end up in an unsafe state, necessitating the use of a safety controller.

\subsubsection{Planner Dynamics and Reward Function} 
\label{sec: Planner dynamics and reward function}
The robot planner state is $\xr = (s_\mathrm{p}, \ell_\mathrm{p}, v_\mathrm{p})$ where $s_\mathrm{p}$ represents the longitudinal distance along a lane, $\ell_\mathrm{p}$ is the lane index, and $v_\mathrm{p}$ is the velocity of the robot. The state representing the $J$ other cars $\xo$ is also of this structure, $\xo = (s_\mathrm{p}^{(1)}, \,\ell_\mathrm{p}^{(1)},\, v_\mathrm{p}^{(1)},\,\ldots,\, s_\mathrm{p}^{(J)}, \,\ell_\mathrm{p}^{(J)},\, v_\mathrm{p}^{(J)})$.
The action space of the planner is $\mathcal{U}=\{$increase $v_\mathrm{p}$ by $1ms^{-1}$, decrease $v_\mathrm{p}$ by $1ms^{-1}$, left lane change, right lane change, idle$\}$. 
Given this state and action representation, we design the following reward function (without the HJI term),
\begin{equation}
\begin{aligned}
    R(\xr, \ur, \uo) = r_\mathrm{speed}(\xr) + r_\mathrm{crash}(\xr, \xo) + r_\mathrm{lane}(\xr)\\
    r_\mathrm{speed}(\xr) = \gamma_1 \frac{v_\mathrm{p} - \underline{v}}{\overline{v} - \underline{v}}, \quad 
    r_\mathrm{lane}(\xr) = \gamma_2 \frac{\ell_\mathrm{p} - \underline{\ell}}{ \overline{\ell} - \underline{\ell}},\\
    r_\mathrm{crash}(\xr, \xo) = -\gamma_3 \mathbf{1} [ \xr \text{ in collision} ],
    \label{eqn:case study reward function}
\end{aligned}
\end{equation}
where $\underline{v}$ and $\overline{v}$ corresponds to the speed limits, $\underline{\ell}$ and $\overline{\ell}$ corresponds to the right-most and left-most lanes on the highway.
We select $\gamma_1 = 0.4$, $\gamma_2=1.0$, $\gamma_3=1.0$, $\underline{v}=15\mathrm{ms}^{-1}$, $\overline{v}=30\mathrm{ms}^{-1}$  for our experiment.
This reward function is designed to encourage the robot to stay in the left most lane, maintain high speed, and avoid collision states. In order to test the effectiveness of our proposed safety controller, we encourage the robot to drive dangerously and weave through dense traffic at a high speed.


\subsubsection{Low-level Controller}
\label{subsubsec:case study low-level controller dynamics}
We use the dynamically extended simple car model for the low-level dynamics of the robot car and all other cars. Let the low-level controller state be $z = (\px,\, \py,\, \theta,\, v)$ and control input be $w = (\delta,\, a)$ (for ease of notation, we drop the subscript r and o denoting robot and the other cars). The equations of motion for the dynamically extended simple car model is, 
\begin{align}
\dot{p}_\mathrm{x} = v \cos{\theta}, \:\:
\dot{p}_\mathrm{y} = v \sin{\theta},\, \:\: 
\dot{\theta} = \frac{ v \tan{\delta}}{L},\:\: 
\dot{v} = a,
\label{eqn:case study low-level dynamics}
\end{align}
where $L$ is the length of the car, $\px$ and $\py$ are the longitudinal and lateral positions of the car in a fixed inertial reference frame respectively, $v$ is the signed speed of the car, and $\delta$ and $a$ are the steering and acceleration commands, respectively. 
For the robot car, a closed-loop feedback controller is deployed to track the desired planner trajectory. In the case of tracking a sequence of waypoints, let $x_{\mathrm{r}}=(s_\mathrm{p}, \ell_\mathrm{p}, v_\mathrm{p})$ be the desired planner state. 
Let $\zr = (\pxr,\, \pyr,\, \thetar,\, \vr)$ be the low-level controller state for the robot and $\Delta \ell$ be the signed lateral distance between the robot and the center line of $\ell_\mathrm{p}$ (left of the centerline is positive). 
The closed-loop feedback controller \cite{Leurent2018} that computes low-level controls to track the desired planner state $\wwr = (\delta,\,\ar)$  is,
\begin{equation}
\begin{aligned}
    \delta = \arctan\left(\frac{-LK_\theta}{\vr}\left[\thetar + \arcsin \frac{K_1 \Delta\ell}{\vr}\right] \right),\: \ar = K_2 (v_\mathrm{p} - \vr)
    \label{eqn:case study closed-loop controller}
\end{aligned}
\end{equation}
where $K_\theta$ , $K_1$, and $K_2$ are control gains to be chosen. In our experiments, $K_\theta=5.0$ , $K_1=2.0$, and $K_2=1.67$.

      
\subsubsection{HJI Relative Dynamics}
To compute the HJI value function for a robot-agent system, we need relative dynamics of the pairwise system.
We use the dynamically extended unicycle model for the robot, and a simplified unicycle model for the other car. 
For ease of notation, we drop the superscript denoting the $j$th pair, but the following equations are referring to a particular robot-agent pair. The dynamics for the robot and other car used for HJI value computations are,
\begin{align}
 z_\mathrm{r, HJI} = \begin{bmatrix}\dot{p}_{x,\mathrm{r}}\\ \dot{p}_{y,\mathrm{r}}\\ \dot{\theta}_\mathrm{r} \\ \dot{v}_\mathrm{r} \end{bmatrix} = \begin{bmatrix} \vr\cos\thetar \\ \vr\sin\thetar \\ \omr \\ \ar \end{bmatrix}, \:\:  z_\mathrm{o, HJI} = \begin{bmatrix}\dot{p}_{x,\mathrm{o}}\\ \dot{p}_{y,\mathrm{o}}\\ \dot{v}_\mathrm{o}  \end{bmatrix} = \begin{bmatrix} \vo\cos\thetao \\ \vo\sin\thetao \\ \ao \end{bmatrix},
    \label{eqn:case study HJI dynamics}
\end{align}
where $w_\mathrm{r, HJI}=(\omr, \ar)$ and $w_\mathrm{o, HJI}=(\thetao, \ao)$ are robot and other car controls, respectively. We assume the control limits: $\underline{\omega}_{\mathrm{r}} \leq \omr \leq  \overline{\omega}_{\mathrm{r}}$, $\underline{a} \leq \ar, \, \ao \leq  \overline{a}$, and $\underline{\theta}_{\mathrm{o}} \leq \thetao \leq  \overline{\theta}_{\mathrm{o}}$. We select dynamics slightly different to \eqref{eqn:case study low-level dynamics} to ensure tractability of \eqref{eqn:HJI multiagent safety controller} (see Section~\ref{subsec:case study HJI safety controller}), and simpler dynamics when modeling the other cars to (i) provide some conservatism to our model by assuming the other cars are more agile than the robot car, and (ii) prevent the relative state from becoming too large since the value function computation suffers from the curse of dimensionality.
We define the relative coordinate frame to be aligned with the inertial frame from \eqref{eqn:case study low-level dynamics} and take the difference in position coordinates, $(\pxrel, \pyrel) = (\pxr - \pxo, \pyr - \pyo)$. Let the relative state between the robot and the $j$th other agent be (still dropping the superscript $j$ for ease of notation) $\xrel = (\pxrel,\, \pyrel,\, \thetar,\, \vr,\, \vo)$. The relative dynamics become,
\begin{equation}
\begin{aligned}
    \dot{p}_\mathrm{x,rel} = \vr\cos\thetar -& \vo\cos\thetao,\quad \dot{p}_\mathrm{y,rel} = \vr\sin\thetar - \vo\sin\thetao \\
    &\dot{\theta}_\mathrm{r} = \omr,\quad \dot{v}_\mathrm{r} = \ar,\quad \dot{v}_\mathrm{o} = \ao
    \label{eqn:case study relative dynamics}
\end{aligned}
\end{equation}

\subsubsection{HJI Value Function}\label{subsubsec:case study HJI value function}

\begin{figure}[t]
    \centering
    \includegraphics[width=0.5\textwidth]{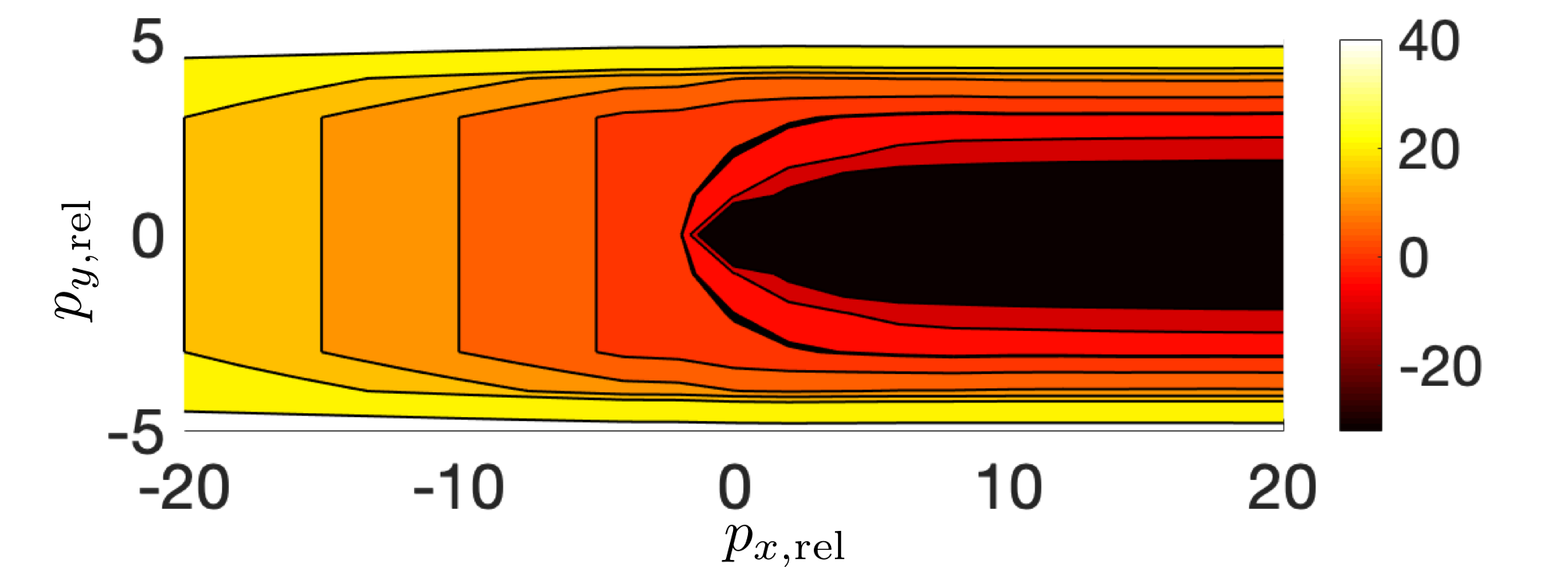}
    \caption{A slice of the HJI value function (see Section~\ref{subsubsec:case study HJI value function}) with $\vo = 20ms^{-1}$, $\vr = 10ms^{-1}$, and the two cars are parallel. $\pxrel >0$ corresponds to the robot car in front of the other car. Since the other car has a higher velocity, it is more unsafe (i.e., negative value) for the robot car to be in front of the other car than behind it.}
    \label{fig:case study hji same ori}
\end{figure}

To compute the value function which represents the set of states the robot wants to avoid,
rather than using the signed distance function to define $V(0,\xrel)$ which is typically used in HJ reachability literature, we instead use the definition of Responsibility-Sensitive Safety (RSS) introduced in \cite{Shalev-ShwartzShammahEtAl2017}. We consider unsafe states to be a function of both relative position, and velocity. For brevity, we refer the reader to \cite{Shalev-ShwartzShammahEtAl2017} for the mathematical formulation, but provide a brief description here.
Safety is decoupled into longitudinal and lateral components.
The longitudinal stopping, $d_\mathrm{long}$, is defined as the distance between a front and rear car if the front car applies maximum braking, and the rear car accelerates maximally over a response time before applying maximum braking.
Analogously, $d_\mathrm{lat}$, is the minimum lateral distance.
A neighboring car is considered unsafe if both the longitudinal and lateral distances between the robot car and that car are less than $d_\mathrm{long}$ and $d_\mathrm{lat}$, respectively. 
Given the velocities corresponding to a particular $\xrel$, we define $V(0,\xrel)$ as follows:
\begin{align}
    V(0, \xrel) = \max( |\pxrel| - d_\mathrm{long},\; 4(|\pyrel| - d_\mathrm{lat})^3).
    \label{eqn:case study V0}
\end{align}
Figure~\ref{fig:case study hji same ori} is a slice of the value function across the $\pxrel$ and $\pyrel$ axes for a time horizon of $|\tau| = 3$ seconds.

\subsection{HJI Safety Controller}
\label{subsec:case study HJI safety controller}

To compute the safety-preserving control set, let $\nabla V(\tau,\xrel) = (\partial V_{\pxrel}, \, \partial V_{\pyrel},\, \partial V_{\thetar},\,\partial V_{\vr}, \partial V_{\vo})$ (for ease of notation, we temporarily drop the superscript $j$, but this is in reference to a particular robot-agent pair). Then,
\begin{align*}
&\min_{\wo \in \mathcal{W}} \nabla V(\tau,\xrel)^T f_\mathrm{rel}(\xrel,\wwr,\wo) = c_1 + c_2 + \partial  V_{\thetar} \omr + \partial V_{\vr} \ar,\\
&c_1 = \min_{\wo \in \mathcal{W}} \:\: \left(\partial V_{\vo} \ao - \partial V_{\pxrel}  \vo\cos\thetao\:  - \partial V_{\pyrel} \vo\sin\thetao \right), \\
&c_2 = \partial V_{\pxrel}  \vr\cos\thetar\:  + \partial V_{\pyrel} \vr\sin\thetar,
\end{align*}
where $c_0 = c_1+c_2$ represents components not dependent on the optimization variables $\wwr$ and $\ar$.
From \eqref{eqn:HJI multiagent safety controller}, the \emph{minimally interventional} multi-agent safety-preserving control is the solution to the optimization problem  (using the superscript notation again for different robot-agent pairs),
\begin{equation}
\begin{aligned}
    \min_{\omr, \ar} &\:\:\: \lambda_1 (\omr - \omega_\mathrm{des})^2 + \lambda_2 (\ar - a_\mathrm{des})^2 + \lambda_3 \max_{k\in\mathcal{K}} \: \eta_k\\
    \mathrm{s.t.} & \:\:\: \partial V_{\thetar}^{(k)} \omr + \partial V_{\vr}^{(k)} \ar \geq -c_0^{(k)} - \eta_k \quad \text{for all } k \in \mathcal{K}\\
    & \:\:\: \eta_k \geq 0 \qquad \qquad \qquad \qquad \qquad \quad\, \text{for all } k \in \mathcal{K}\\
    & \:\:\: \underline{\omega}_{\mathrm{r}} \leq \omr \leq  \overline{\omega}_{\mathrm{r}},\qquad  \underline{a}_{\mathrm{r}} \leq \ar \leq  \overline{a}_{\mathrm{r}}.
\end{aligned}
\label{eqn:case study multiagent safety controller}
\end{equation}
where $\lambda_i>0$ are weights 
(we use $\lambda_1= \overline{\omega}_r^{-2}, \lambda_2= \overline{a}^{-2}, \lambda_3= 1$), $\eta_k$ are slack variables, and $\omega_\mathrm{des}$ and $a_\mathrm{des}$ are the desired controls from \eqref{eqn:case study closed-loop controller} (to map the HJI safety control $\omega$ to $\delta$, we use $\delta =  \tan^{-1}\frac{\omega L}{v}$). \eqref{eqn:case study multiagent safety controller} is minimally interventional because it minimizes tracking error subjected to safety constraints.
Alternatively, the robot can use a switching strategy by letting $\lambda_2 = 0$, $\omega_\mathrm{des} = \omega_\mathrm{prev}$ (to discourage discontinuous steering inputs), and removing the $\eta_k \geq 0$ constraint.
The safety controller will choose controls that satisfy, or violate, each safety constraint equally (i.e., it prioritizes safety for all cars equally).
By our choice of HJI dynamics \eqref{eqn:case study HJI dynamics}, the optimization problem~\eqref{eqn:case study multiagent safety controller} is convex and we solve it using CVXPY \cite{DiamondBoyd2016}.


\subsection{Experimental Results}

\begin{table*}[t]
\centering
\begin{tabular}{|ccc|ccccccccc|}
\hline
Planner & Controller    & Scheme & 
\begin{tabular}{@{}c@{}} TTC \\ $\ge 3$ \end{tabular} &
\begin{tabular}{@{}c@{}} TTC $10^{th}$ \\ percentile \end{tabular} &
\begin{tabular}{@{}c@{}} BTN \\ $\le 1$ \end{tabular} & 
\begin{tabular}{@{}c@{}} BTN $90^{th}$ \\ percentile \end{tabular} &
\begin{tabular}{@{}c@{}} STN \\ $\le 1$ \end{tabular} & 
\begin{tabular}{@{}c@{}} STN $90^{th}$ \\ percentile \end{tabular} & 
\begin{tabular}{@{}c@{}}Mean $\vr$ \\ ($ms^{-1}$)\end{tabular}  & 
\begin{tabular}{@{}c@{}}Mean $|\ar|$ \\ ($ms^{-2}$)\end{tabular}   & 
\begin{tabular}{@{}c@{}} Interven- \\ tions \% \end{tabular} \\\hline

\multicolumn{3}{|c|}{$\uparrow$: higher is better, $\downarrow$: lower is better} & $\uparrow$ & $\uparrow$ & $\uparrow$ & $\downarrow$ & $\uparrow$ & $\downarrow$ & $\uparrow$ & $\downarrow$ & - \\
\hline
HJOP   & SPC        & SW     & 1.000 & 9.927 & 1.000 & 0.079 & 1.000 & 0.016 & 21.878 & 1.273 & 9.5 \\

HJOP   & RSS           & SW     & 0.996 & 17.125 & 0.998 & 0.030 & 0.994 & 0.006 & 20.343 & 4.066 & 55.0 \\

\textbf{HJOP}   & \textbf{SPC} & \textbf{MI} & 0.999& 9.607 & 0.995& 0.109 & 0.994& 0.017 & 22.000& 1.154 & 18.1\\
HJOP & RSS          & MI     & 0.996 & 13.914 & 0.998 & 0.028 & 0.995 & 0.008 & 20.301 & 2.788 & 57.5 \\

HJOP   & None          & ---    & 0.956 & 7.009 & 0.975 & 0.213 & 0.982 & 0.034 & 22.554 & 0.416 & 0.0 \\
 \hline
OP      & SPC        & SW     & 0.996 & 7.045 & 0.977 & 0.152 & 0.960 & 0.084 & 21.144 & 5.940 & 51.7 \\

OP      & RSS           & SW     & 0.998 & 13.148 & 0.999 & 0.030 & 0.998 & 0.009 & 19.571 & 5.398 & 71.2 \\

OP      & SPC        & MI     & 0.999 & 8.417 & 0.988 & 0.091 & 0.984 & 0.040 & 21.039 & 3.470 & 63.2 \\

OP      & RSS           & MI     & 0.997 & 13.971 & 1.000 & 0.024 & 0.998 & 0.008 & 21.074 & 2.596 & 67.2 \\

OP      & None          & ---    & 0.502 & 0.785 & 0.714 & 5.616 & 0.782 & 3.189 & 28.141 & 0.386 & 0.0 \\

\hline
\end{tabular}
\caption{Statistics for different planner and safety controller configurations. The values in the $\mathrm{TTC}\geq 3$, $\mathrm{BTN}\leq 1$, and $\mathrm{STN}\leq 1$ column represent the fraction of samples that satisfy the inequality. Our proposed method is in bold.}
\label{tab:case study results statistics}
\end{table*}

To evaluate the benefits of our proposed planning and control stack, 
we perform an ablation study and compare our approach to the RSS policy proposed in \cite{Shalev-ShwartzShammahEtAl2017}. For the high-level planner, we investigate the following configurations,
\begin{itemize}
    \item \textbf{OP} An OP planner only, i.e., $\gamma_R = 1$ in \eqref{eqn: HJI planner reward function}.
    \item \textbf{HJOP} An OP planner with a HJI reward term, i.e., $\gamma_R = 0.9$ in \eqref{eqn: HJI planner reward function}.
\end{itemize}
With a fixed planner (HJOP or OP), we investigate different low-level safety control strategies used either in a switching (\textbf{SW}), or minimally interventional (\textbf{MI}) scheme (see Section~\ref{subsec:case study HJI safety controller} for the formulation):

\begin{itemize}
    \item \textbf{None} No safety controller is used.
    \item \textbf{RSS} The RSS proper response policy proposed in \cite{Shalev-ShwartzShammahEtAl2017} provides the minimum longitudinal and lateral acceleration necessary to maintain safety.
    We use the RSS proper response set instead of the HJI constraint in \eqref{eqn:case study multiagent safety controller}.
    Under this RSS framework, there will be no accidents where the autonomous vehicle is at fault from a planning perspective.

    \item \textbf{SPC} Our proposed multi-agent HJI safety-preserving controller (SPC) in \eqref{eqn:case study multiagent safety controller}.

\end{itemize}
To compare each approach, we use the following metrics, 
\begin{itemize}
    \item \emph{Time-to-Collision (TTC):}  The estimated time before collision assuming both vehicles continue at constant speed \cite{Jansson2005}. Lower values indicate a more dangerous situation, while higher values are better with diminishing return.
    \item \emph{Brake Threat Number (BTN)} and \emph{Steer Threat Number (STN): }The required longitudinal and lateral acceleration for collision avoidance as defined in \cite{BraennstroemSjoebergEtAl2008} divided by  maximum available longitudinal/lateral acceleration. Lower values indicate safer situations.
    \item \emph{Mean velocity:} The mean speed of the robot car over all sample points. Higher values indicate better performance (based on the planner's reward function).
    \item \emph{Mean acceleration magnitude:} The mean magnitude of acceleration of the robot car over all sample points. Lower values indicate more efficient driving.
    \item \emph{Intervention percentage:} Percentage of samples where the safety controller stepped in. Lower values implies a safer planner.
\end{itemize}

\begin{figure}[t]
    \centering
    \includegraphics[width=0.45\textwidth]{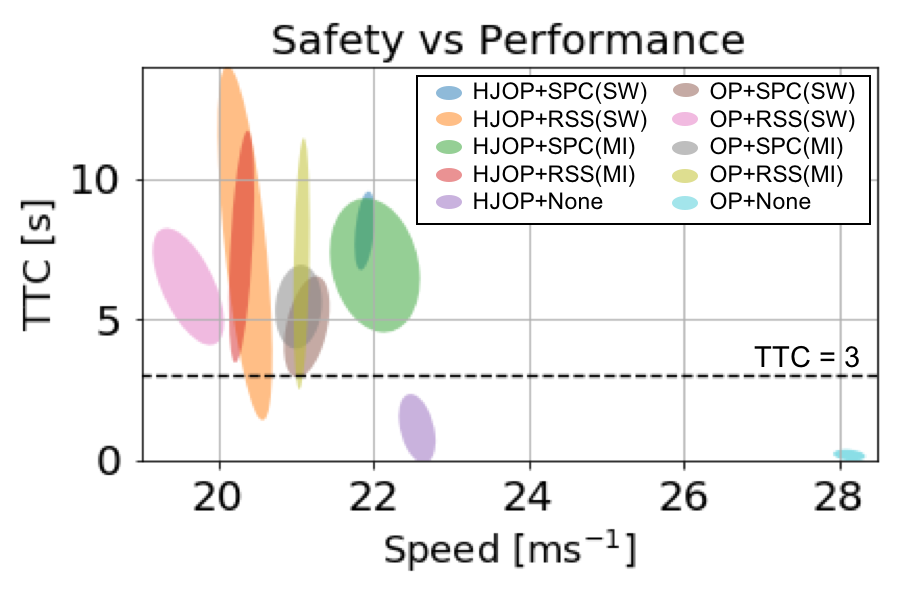}
    \caption{The trade-off between safety (1$^{st}$ percentile TTC) and performance (mean speed) computed from samples taken over ten-second intervals across each episode. Three standard deviation ellipses are shown.}
    \label{fig:case study safety vs efficiency}
\end{figure}
We performed 20 episodes of 30-second highway simulation with 50 Hz data collection for each planner-controller configuration. Each episode consists of 100 other vehicles that the robot needs to weave through as shown in Figure \ref{fig:highway env}. This set up provides a sufficiently dense and challenging environment to evaluate the effectiveness of each planner-controller configuration.
Statistics of the simulations are listed in Table~\ref{tab:case study results statistics}, and the trade-off between safety, measured by the 1$^{st}$ percentile TTC, and performance, measured by mean speed, is shown in Figure~\ref{fig:case study safety vs efficiency}. We select $\mathrm{TTC}=3$ as the boundary between unsafe and safe to reflect the popular ``three-second rule'' while driving, but note that this threshold could be different.

We highlight three key takeaways from these results. 
First, in the absence of a safety controller, adding an additional HJI term to the planner's reward function significantly improves safety as measured by BTN, STN and TTC, but, as expected, with a decrease in performance (i.e., mean speed). We note that the efficiency (i.e., mean magnitude of acceleration) of these two configurations are similar. 
Second, SPC and RSS provide very similar levels of safety assurance (considering the diminishing returns of larger TTC), yet SPC provides better performance and higher efficiency than RSS. 
RSS experiences more interventions, and therefore is likely to execute more braking and swerving maneuvers. We hypothesize that when using the RSS safety controller, the misalignment of the notion of safety between the planner and RSS controller results in chattering. 
Third, the MI scheme is more efficient (i.e., lower acceleration) than SW, yet performs similarly well in terms of safety metrics. 
In practice, using MI may be more desirable because the SPC and RSS controller assumes worst-case outcomes by the other agents, but this may not necessarily be what happens over the entire interaction. In other words, MI prevents the robot from \emph{overreacting} every time safety is nearly violated.
Instead of prioritizing all agents equally, future work can be directed at assigning priority to which agent to avoid. For example, weighing each safety constraint based each agent's likelihood of following an adversarial policy.

Further, we can visualize the safety-performance trade-off in Figure~\ref{fig:case study safety vs efficiency}.
The OP+None configuration is clearly the fastest but the most unsafe.
The top right corner corresponds to the region with highest safety and performance. Although the RSS controllers with HJOP provides higher safety metrics, the diminishing return nature of TTC indicates that our proposed method HJOP+SPC (green) is in the ideal region---it is above the $\mathrm{TTC}=3$ line, and furthest to the right.

\subsection{The Mechanism Behind HJOP+SPC}
\begin{figure}[t]
    \centering
    \includegraphics[width=0.48\textwidth]{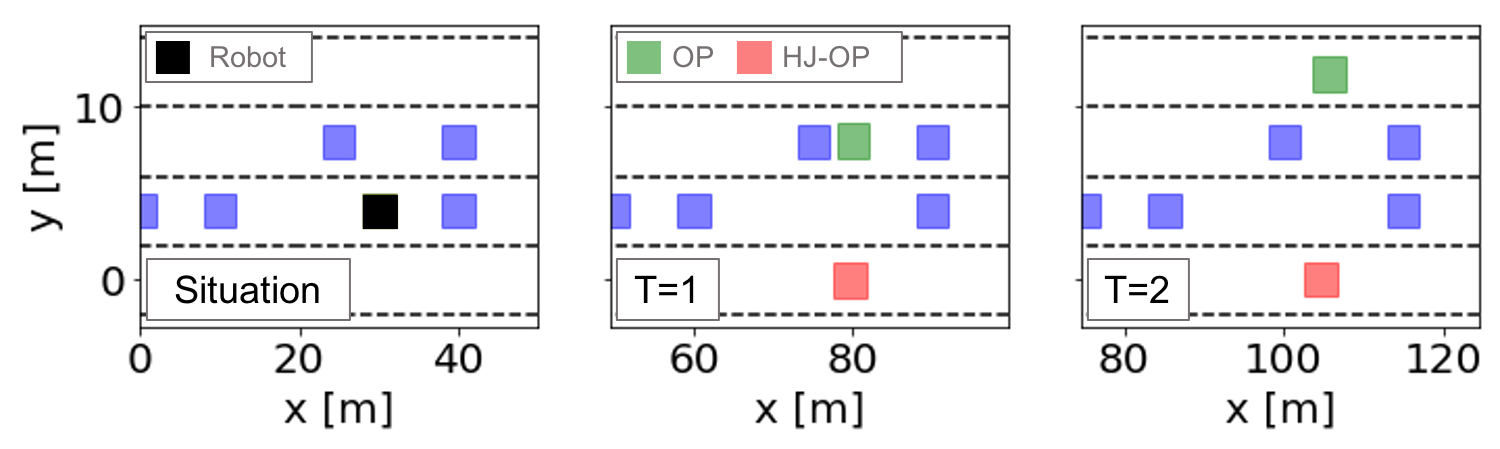}
    \caption{Snapshots of the robot's desired plan using different planning strategies.\\}
    \label{fig:case study planner comparison snapshots}
\end{figure}

\begin{figure}[t]
    \centering
    \includegraphics[width=0.48\textwidth]{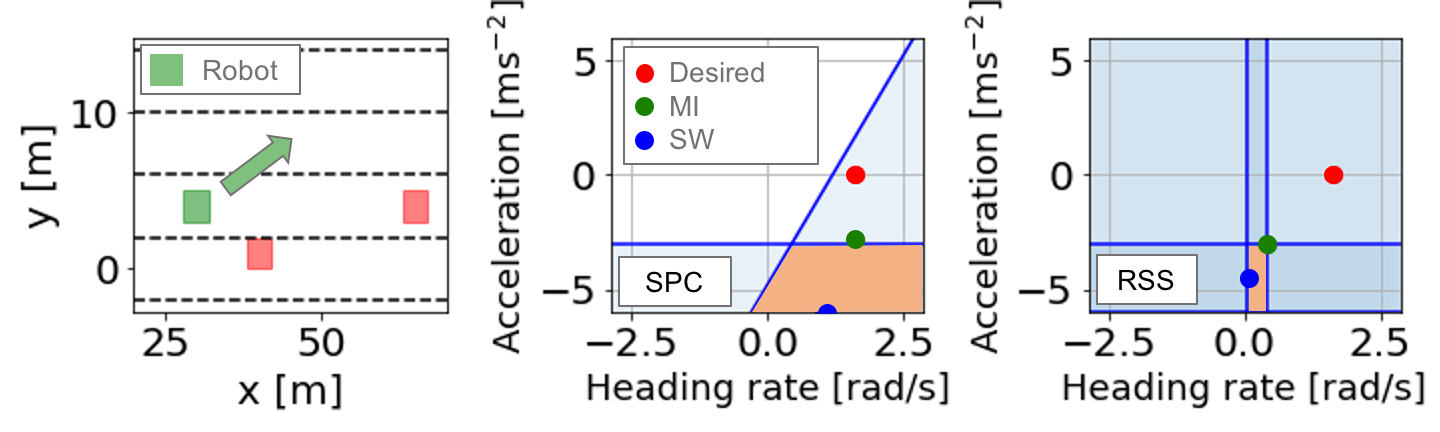}
    \caption{Optimal controls using the SPC and RSS safety controller. In this example, $\omega_\mathrm{prev}=0$. Left: Front red car is traveling 20ms$^{-1}$, and other two cars are traveling at 25ms$^{-1}$.}
    \label{fig:case study controller comparison plots}
\end{figure}

We take a closer look at how the HJOP and OP planners behave in a potentially dangerous situation. Consider the situation shown in Figure~\ref{fig:case study planner comparison snapshots} (left); the desired plan computed by each planner which, for this example only, assumes the other cars move at constant speed. The next two plots show the robot's roll-out over the next two time steps. The OP planner (in green) chooses to squeeze in between the two cars on its left, while the HJOP planner (in red) changes to the right lane to avoid being trapped by the other cars.
Figure~\ref{fig:case study controller comparison plots} visualizes the set of safety preserving controls considered by the SPC and RSS controllers for a situation where the robot (in green) desires to move to the left lane (see left figure) but safety is near violation by two agents. We see the safety-preserving control constraints imposed by HJ reachability and RSS (orange region indicates the intersection of all safety-preserving control sets), and the solution to \eqref{eqn:case study multiagent safety controller} for the MI and SW cases. The box-constraint nature of RSS makes it more restrictive than SPC. Due to linear constraints, MI chooses controls on the boundary and as close to the desired control as possible. While SW selects controls close to the $\omega_\mathrm{prev}$ while satisfying the control constraints as much as possible.


\section{Conclusions and Future Work}
\label{sec: conclusion}
By sharing the same interpretation of what it means to be safe, the planning and control modules which are typically designed separately are now complementary and can provide safe yet performant behaviors. 
Our proposed approach leverages HJ reachability theory and, as demonstrated with a multi-agent highway driving scenario, equips a robot with the foresight to avoid regions of possible inevitable collision, and the ability to minimally deviate from the desired trajectory to the extent necessary to avoid collision with multiple agents.
We propose three future research directions; the first would be to extend the idea of using a HJI value function as part of the objective function beyond planning algorithms, such as for trajectory optimization, or as a feature in inverse reinforcement learning. The second would entail exploring different ways to define the target set (i.e., $V(0,x)$) when computing the value function, and understanding how it affects the safety-performance trade-off.
The third is to design smarter priority assignment, such as through chance constraints, when safety is nearly violated by multiple agents.

\bibliographystyle{unsrt} 

\bibliography{main,ASL_papers}

\end{document}